\documentclass{article}

     \usepackage[nonatbib,preprint]{neurips_2019}

\usepackage[utf8]{inputenc} 
\usepackage[T1]{fontenc}    
\usepackage{hyperref}       
\usepackage{url}            
\usepackage{booktabs}       
\usepackage{amsfonts}       
\usepackage{nicefrac}       
\usepackage{microtype}      

\usepackage{graphicx}       
\usepackage{wrapfig}

\usepackage[ruled]{algorithm2e}
\usepackage{algorithmic}
\usepackage{amsmath}
\usepackage{enumitem}
\usepackage{xcolor}

\usepackage[labelfont=bf]{caption}
\title{Learning to Disentangle Robust and Vulnerable Features for Adversarial Detection}

\author{%
  Byunggill Joe \\
  KAIST \\
  \texttt{cp4419@kaist.ac.kr} \And
  Sung Ju Hwang  \\
  KAIST \\
  \texttt{sjhwang82@kaist.ac.kr} \And
  Insik Shin \\
  KAIST \\
  \texttt{insik.shin@kaist.ac.kr}
}

\begin{document}

\maketitle

\begin{abstract}
Although deep neural networks have shown promising performances on various tasks, even achieving human-level performance on some, they are shown to be susceptible to incorrect predictions even with imperceptibly small perturbations to an input. There exists a large number of previous works which proposed to defend against such adversarial attacks either by robust inference or detection of adversarial inputs. Yet, most of them cannot effectively defend against whitebox attacks where an adversary has a knowledge of the model and defense. More importantly, they do not provide a convincing reason why the generated adversarial inputs successfully fool the target models. To address these shortcomings of the existing approaches, we hypothesize that the adversarial inputs are tied to \emph{latent} features that are susceptible to adversarial perturbation, which we call \textit{vulnerable} features. Then based on this intuition, we propose a minimax game formulation to disentangle the latent features of each instance into robust and vulnerable ones, using variational autoencoders with two latent spaces. We thoroughly validate our model for both blackbox and whitebox attacks on MNIST, Fashion MNIST5, and Cat \& Dog datasets, whose results show that the adversarial inputs cannot bypass our detector without changing its semantics, in which case the attack has failed.
\end{abstract}


\section{Introduction}
Although deep neural networks have achieved impressive performances on many tasks, sometimes even surpassing human performance, researchers have found that they could be easily fooled by even slight perturbations of inputs. Adversarial examples, which are deliberately generated to change the output without inducing semantic changes in the perspective of human perception~\cite{szegedy,fgsm}, can sometimes bring down the accuracy of the model to zero percent. Many of previous work try to solve this problem in the forms of robust inference~\cite{dae,pgd,stability,defensegan,towardmnist,sap}, which aims to obtain correct results even with the adversarial inputs, or by detecting adversarial inputs~\cite{nic,kd-detect,gong-detect,grosse-detect,rce,feature-squeeze,mutation-detect}. 

However most of previous works do not work properly in whitebox attack scenarios where the adversary has the same knowledge as a defender. Most of the past defenses which at a time successfully defended against adversarial attacks were later broken. For example, adversarial defenses leveraging randomness, obfuscated gradients, input denoising, and neuron activations which were once deemed as robust, were later broken with sophisticated whitebox attacks such as expectation over transformation or backward pass differentiable approximation~\cite{obfuscated, easily}. 

\setlength{\columnsep}{0pt}%
\begin{wrapfigure}{R}{0.4\textwidth}
\vspace{0.0\baselineskip}
   \begin{minipage}{0.4\textwidth}
 \resizebox{1.0\textwidth}{!}{
  \centering
       \includegraphics[scale=0.2]{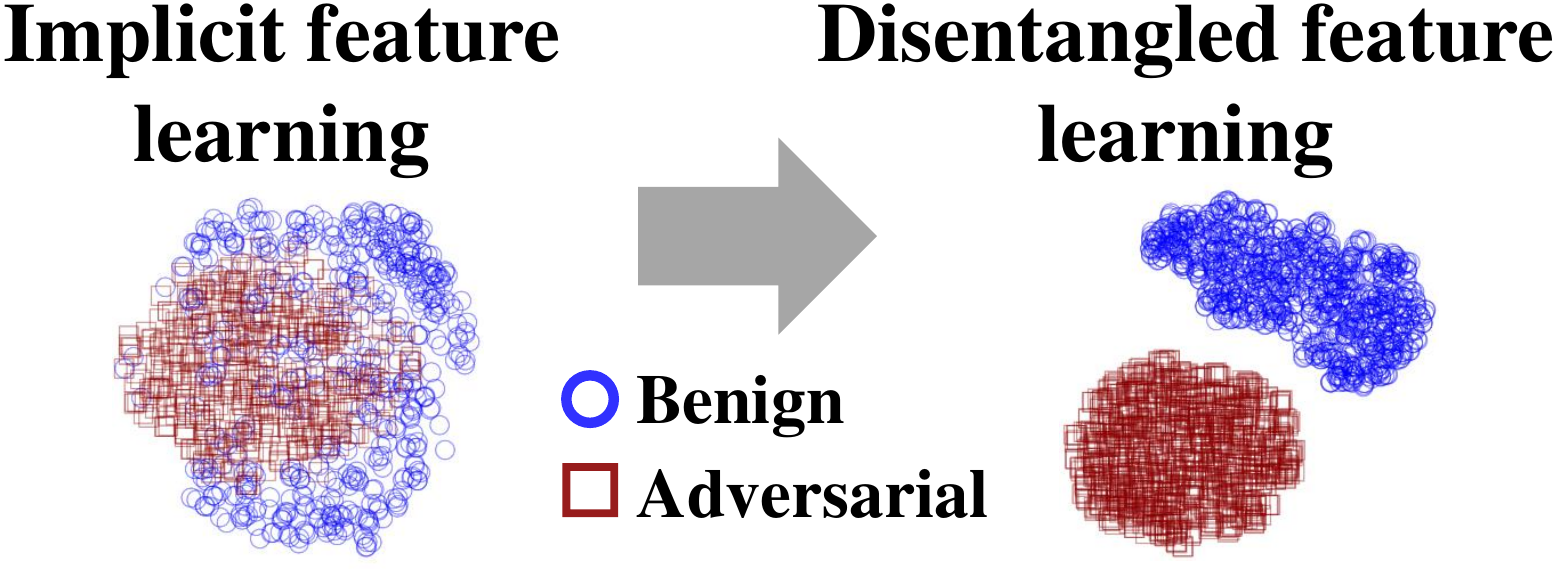}
    }                                        
\vspace{-0.9\baselineskip}
\caption{Latent spaces of features}
\vspace{-0.9\baselineskip}
   \end{minipage}
\end{wrapfigure}
\label{fig:entangled}

We hypothesize that the malfunctions of the previous defenses are caused by the existence of features that are more susceptible to adversarial perturbations, which misguide the model to make incorrect predictions. Such a concept of vulnerable features has been addressed in a few existing works~\cite{odd,nrf}. These approaches define vulnerability of the \emph{input features} either based on the human perception, based on the amount of perturbation at the network output, or correlation with the label. Yet, our hypothesis has a different perspective of vulnerability/robustness in that we assume that there exists a latent feature space where adversarial inputs to a label form a common latent distribution. That is, vulnerable and robust features by our definition are not input features, but rather \emph{latent features} residing in hypothetical spaces. While vulnerability/robustness of features are assumed as given and fixed in the existing work, this redefinition allows us to explicitly \emph{learn} to disentangle features into robust and vulnerable features by learning their feature spaces. 

Toward this goal, we propose a variational autoencoder with two latent feature spaces, for robust and vulnerable features respectively. Then, we train this model using a two-player mini-max game, where the adversary tries to maximize the probability that the adversarially perturbed instances are in the robust feature space, and the defender tries to minimize this probability. This procedure of \emph{learning} to disentangle robust and vulnerable features further allows us to \emph{detect} features based on their likelihood of the feature belonging to either of the two feature categories (see~\autoref{fig:entangled}).

We validate our model on multiple datasets, namely MNIST~\cite{mnist}, Fashion MNIST5~\cite{fashion-mnist}, and Cat \& Dog~\cite{cat-and-dog}, and show that the attacks cannot bypass our detector without incurring semantic changes to the input images, in which the attack has failed. 

Our contributions can be summarized as follows:

\begin{enumerate}
\item We empirically show that adversarial attacks are negative side effects of \emph{vulnerable features}, which is the byproduct of \emph{implicit representation learning} algorithms.
\item From the above empirical observation, we hypothesize that different adversarial inputs to a label form a common latent distribution.
\item Based on this hypothesis, we propose a new defense mechanism based on variational autoencoders with two latent spaces, and a two-player mini-max game to learn the two latent spaces, for robust and vulnerable features each, and use it as an adversarial input detector.
\item We conduct blackbox and whitebox attacks to our detector and show that adversarial examples cannot bypass it without inducing semantic changes, which means attack failure.
\end{enumerate}

%

\section{Related work}
After revealing severe defects of neural networks
against adversarial inputs\cite{szegedy, fgsm},
researchers have found more sophisticated attacks to
fool the neural networks\cite{fgsm,pgd,mim,cw,deepfool,elasticnet}. 
On the other side, many researchers have tried to propose
a defense mechanism against such attacks as a form of
robust prediction~\cite{dae,pgd,stability,defensegan,towardmnist,sap}
or detector of adversarial inputs~\cite{nic,kd-detect,gong-detect,grosse-detect,rce,feature-squeeze,mutation-detect}.
Many of them rely on randomness, obfuscated gradients,
or distribution of neuron activations, which are known to malfunction
under adaptive whitebox attacks~\cite{obfuscated, easily}.
Some of previous works studied root causes of this intriguing
defects. Goodfellow et al.\cite{fgsm} suggest there should be vulnerability
in benign input distributions with an interpretation of deep neural networks
as linear classifiers.
Tsipras et al.\cite{odd} further propose a dichotomy of robust feature
and non-robust feature. They analyze the intrinsic
trade-off between robustness and accuracy.
Concurrent to this work, Ilyas et al.\cite{nrf} find that
non-robust features suffice for achieving good accuracy
on benign inputs. They also provide a theoretical framework 
to analyze the non-robust features. On the other hand,
we provide a hypothesis about \emph{latent distributions of
vulnerable features}, and disentangle
vulnerable latent space from the entire latent feature space
under adaptive whitebox attacks.


\section{Background and Motivational Experiment}

\paragraph{Premise.} The key premise of our proposed framework is that standard neural-network classifiers implicitly learn two types of features: \textit{robust} and \textit{vulnerable} features. The robust features, in an intuitive sense, correspond to signals that make semantic sense to humans, which may describe texture, colors, local shapes or patches in image domain. The vulnerable features are considered as imperceptible to human senses but are leveraged by models for prediction since they help lowering the training loss. We posit that adversarial attacks exploit vulnerable features to imperceptibly perturb inputs that induce erroneous prediction.
\paragraph{Notation.} We now present notations used throughout this paper. Firstly, $x \in \rm I\!R^{d}$ is an input in $d$ dimensions.
We distinguish $x$ between benign input $x_{b}$ and adversarial input $x_{a}=x_{b} + \delta$,
where $\delta \in S=\{\delta \mid||\delta||_p < \epsilon\}.$
Typical $p$ is one of $0, 1, 2,$ and $\infty$.
We use an ordered set $X$ with matching subscripts $b$ and $a$
to indicate a set of $x_b$ and $x_a$ and with superscript $c$ to indicate label $c$.
Given an input $x_b$, the true label of $x_b$ is $y_b$,
and $y_a$ is an inaccurate (misled) label of corresponding $x_a$.
$Y_b$ and $Y_a$ are ordered sets of labels of $x_b$ and $x_a$
that have the same indice in $X_b$ and $X_a$.
The number of unique labels in classification is $N$.
We denote a dataset as a pair ($X$, $Y$).
$P^c_r$ and $P^c_v$ are the probatility density functions of robust features and vulnerable features respectively for the inputs of label $c$.
$\theta^c$ is a set of model parameters of a variational autoencoder for label $c$ (i.e., VAE$^{c}$).
\paragraph{Attack methods.}
 We briefly explain representative attack methods used in this paper.
 
\textbf{Fast Gradient Sign Method (FGSM)} is an one step method proposed by Goodfellow et al.~\cite{fgsm}. 
$L(x,y)$ is a training loss. The attack takes sign of $\nabla L(x,y)$ and perturbs
the $x$ with a size parameter $\epsilon$ to increase $L(x,y)$, resulting in an unexpected result.
    $$x_{a} = x + \epsilon \cdot sign(\nabla L(x, y))$$
\textbf{Projected Gradient Decent (PGD)}
is an iterative version of the FGSM attack with random start~\cite{pgd}.
To generate $x_a^{t+1}$, the attack perturbs $x_a^{t}$ with a step size parameter $\alpha$
based on the sign of $\nabla L(x^t_a,y)$. It limits its search space in the input space of $x+S$,
which is implemented as a clipping function with a $L_\infty$ bound.
It can initialize the attack, adding uniform noise to an original image, $x^0_a=x + U(S)$.
    $$x_{a}^{t+1} = \prod_{x+S}(x_{a}^{t} + \alpha \cdot sign(\nabla L(x_{a}^{t}, y)))$$
\textbf{Momentum Iterative Method (MIM)}
introduces concept of momentum to the PGD attack~\cite{mim}.
Instead of directly updating $x^t_{a}$ from $\nabla L(x^{t},y)$, it applies previous $g^{t}$ value with
decay factor $\mu$ to $g^{t+1}$. Then it updates $x_a^{t+1}$ with the sign of $g^{t+1}$ and the step size
$\alpha$, while limiting search space in $x+S$.
    $$g^{t+1} = \mu \cdot g^{t} + \frac{\nabla {L(x_a^{t}, y)}}{||\nabla {L(x_a^{t}, y)}||_1},\;\;x_{a}^{t+1} = \prod_{x+S}(x_{a}^{t} + \alpha \cdot sign(g^{t+1})) $$
\textbf{Carlini \& Wagner Method (CW-L2)}~\cite{cw} introduces $w$ and searches
adversarial inputs on the $w$ space, because it relaxes
discontinuous property at the minimum and maximum input values.
To minimize the distortion of adversarial inputs, it incorporates a $L_2$ loss term,
$||\frac{1}{2}(\mathrm{tanh}(w)+1) - x||_2^2$.
It defines a function $f(x')$ to induce miss-classification, where $Z(x')_i$ is the logit value of label $i$ right before a softmax layer, and $t$ is a label of the original input $x$.
The $\kappa$ is for larger differences in logit values, resulting in high confidence.
It balances between $L_2$ loss and $f(x')$ loss in a binary search of $c$.
    $$ min_{w}||\frac{1}{2}(\mathrm{tanh}(w)+1) - x||_2^2 + c \cdot f(\mathrm{tanh}(w) + 1),\;\;f(x') = max(max(Z(x')_i:i\neq t) - Z(x')_t, -\kappa)$$

\footnotetext{The sign of D loss depends whether an attack uses gradient descent(+) or ascent(-).}

\subsection{Vulnerability of Implicit Feature Learning} \label{sec:motivational}

By showing that the vulnerable features are prevalent in benign datasets ($X_b$, $Y_b$), we validate our argument that the implicit learning algorithms may lead models to learn vulnerable features that could be exploited by an adversary, without explicit regularizations to prevent learning of them. We now provide an evidence in support of this, where a benign dataset ($X_b$, $Y_b$) can be classified with good accuracy by a classifier $F_v$ which is trained with a dataset ($X_v$,$Y_v$) only containing vulnerable features.

\begin{figure}
  \centering
    \includegraphics[scale=0.68]{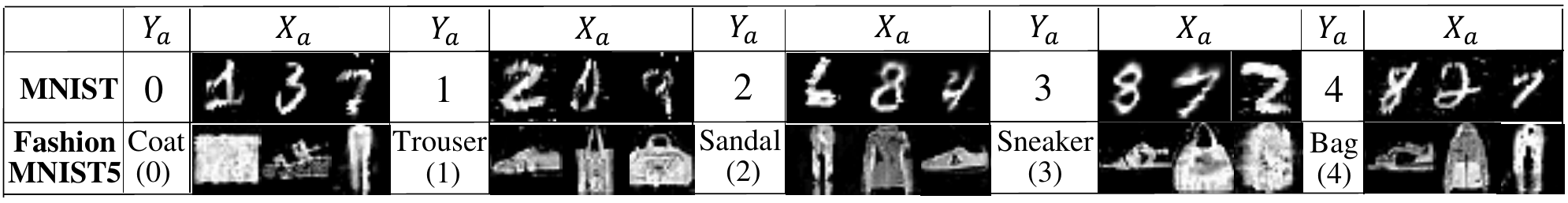}
\vspace{-0.4\baselineskip}
  \caption{Examples of vulnerable feature dataset ($X_v$, $Y_v$) generated by the CW-L2\_d attack}
\vspace{-0.8\baselineskip}
  \label{fig:vfr-training-example}
\end{figure}

The basic idea to construct ($X_v$, $Y_v$) for $F_v$ is to leverage a dataset of adversarial inputs ($X_a$,$Y_a$) that an arbitrary attack $A$ generates, to fool a classifier $F_p$ pre-trained with ($X_b$, $Y_b$), where $Y_a$ is a set of incorrect labels. The set of adversarial inputs $X_a$ cause $F$ to make erroneous prediction toward a target label $y$. That is, each $X_a$ contains a set of vulnerable features that $F_p$ learned to identify $y$, because the features do not contain any semantically meaningful features of $y$ (see \autoref{fig:vfr-training-example}). Based on this reasoning, we construct ($X_v$, $Y_v$) by attacking $F_p$ and accumulating a set of $X_a$ with the inaccurate labels $Y_a$. For clarity, ($X_v$, $Y_v$) should not contain robust features; it should not include benign inputs and should not allow large distortions that could change the semantics of the inputs.

\begin{table}[t]
  \scriptsize
  \centering
    \caption{Performance of a classifier $F_v$ trained only with vulnerable feature dataset ($X_v$, $Y_v$)}
    \resizebox{\textwidth}{!}{
\begin{tabular} {l c c c c c c c c}

\toprule
                          & \multicolumn{4}{c}{\textbf{MNIST}}                                                   & \multicolumn{4}{c}{\textbf{Fashion MNIST5}}                                \\
                          \cmidrule(r){2-5}                                                             \cmidrule(r){6-9}
                          & \textbf{FGSM\_d} & \textbf{PGD\_d} & \textbf{MIM\_d}  & \textbf{CW-L2\_d}   & \textbf{FGSM\_d} & \textbf{PGD\_d} & \textbf{MIM\_d}     & \textbf{CW-L2\_d} \\
\midrule
    Test $X_b$ accuracy        &0.37              &0.85             &0.92              &\textbf{0.98}        &0.53              &0.71             &\textbf{0.78}        &\textbf{0.78}      \\
    Test $X_v$ accuracy        &0.54              &0.81             &0.73              &0.96                 &1.0               &0.95             &0.93                 &1.0                \\
    \# of iterations      &2                 &7                &30                &30                   &3                 &30               &30                   &30                 \\
    $D$ loss term of $A$\footnotemark& $-100D(x)$       &$-10D(x)$       &$-2D(x)$          &$0.5D(x)$             &$-10D(x)$         &$-10D(x)$        &$-1D(x)$             &$0.5D(x)$          \\
\bottomrule 
\end{tabular}

%


    }
    \label{table:intro:vfr-training}
  \vspace{-2em}
\end{table}

To make the accumulation process efficient, we introduce a discriminator $D$ which learns the vulnerable features in ($X_v$, $Y_v$) and prevents the attack $A$ from re-exploiting the vulnerable features that has already learned. The attack $A$ should then bypass $D$, which is only possible with exploiting a new set of vulnerable features. After a pre-determined number of iterations for accumulation, we train $F_v$ with $(X_v, Y_v)$ and measure the performance of $F_v$ on $(X_b, Y_b)$ and test datasets $(X'_v, Y'_v)$.


We conduct this experiment \footnote{It is worth noting that a similar experiment~\cite{nrf} was conducted independently at the same time. A key difference is that the experiment in~\cite{nrf} is designed to construct \textit{non-robust} (vulnerable) input datasets while our experiment aims to construct a common latent distribution of a maximal set of vulnerable features with $D$.} with two datasets, MNIST and Fashion MNIST5. In the case of Fashion MNIST5, we use a subset of Fashion MNIST~\cite{fashion-mnist} which are "Coat (0)", "Trouser (1)", "Sandal (2)", "Sneaker (3)", and "Bag (4)".  
\autoref{fig:vfr-training-example} illustrates vulnerable feature datasets ($X_v$, $Y_v$) generated by $A$. 
We can see that $X_v$ describes a completely different visual object classes from $Y_v$, as $X_v$ is comprised of only the vulnerable features.
For more details of the experiment, see \autoref{appendix:expvfr} in \textbf{supplementary file}.

We use four representative attacks, including FGSM\_d, PGD\_d, MIM\_d, and CW-L2\_d, where "\_d" indicates that each individual base attack is adapted to bypass $D$ with an additional attack objective.
\autoref{table:intro:vfr-training} summarizes the performance of $F_v$. Interestingly, the results show that it is possible to achieve high accuracy of up to 0.98 (MNIST) and 0.78 (Fashion MNIST5) on ($X_b$, $Y_b$), even though they are trained with ($X_v$, $Y_v$). We note that $F_v$ also achieves high accuracy on test datasets ($X'_v$, $Y'_v$), which are unseen in the training phase.


From the above experiments, we can draw the following two conclusions:
\begin{itemize} 
    \item High accuracy on ($X_b$, $Y_b$): Vulnerable features are prevalent in ($X_b$, $Y_b$), and we need new training algorithms that are able to distinguish vulnerable and robust features for robustness.
    \item High accuracy on ($X'_v$, $Y'_v$): ($X_v$, $Y_v$) must share some high-level features in common, although they may be imperceptible to humans. Based on this observation, we hypothesize that the adversarial inputs $X_v$ of the same label $Y_v$ exist in the common latent distribution. \\
\end{itemize}
\vspace{-1.2\baselineskip}

\section{Approach}
\begin{figure}[h]
  \centering
  \includegraphics[scale=0.55]{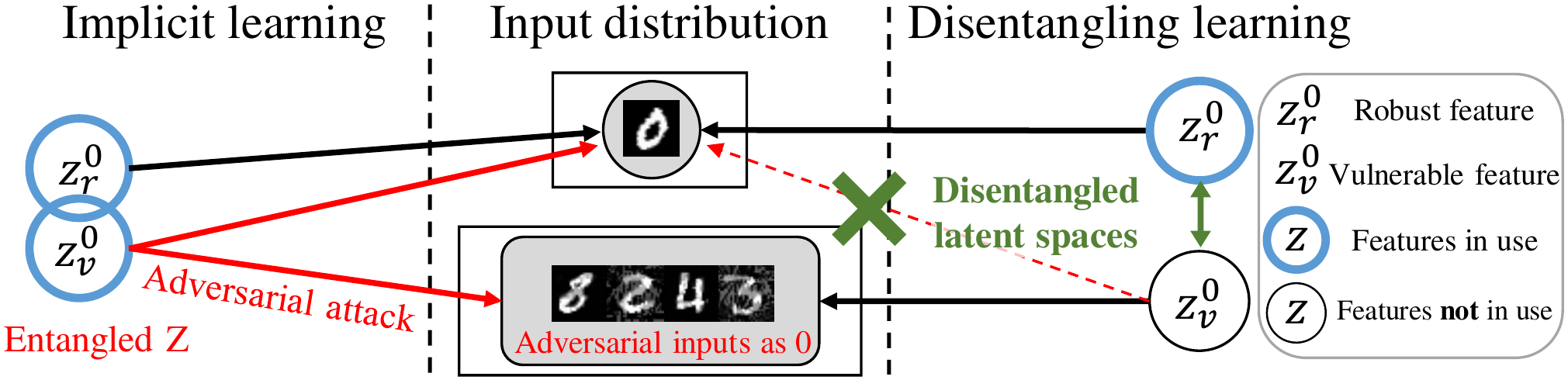}
  \caption{Implicit vs. disentangling learning in latent feature spaces (label 0 on MNIST)}
  \label{fig:vfr-illustration}
\end{figure}

Based on the hypothesis that vulnerable features make up a common latent distribution for each label, we propose a new learning algorithm to recognize and disentangle a latent space of the vulnerable features from distributions of the all features. \autoref{fig:vfr-illustration} shows the difference from an implicit algorithm (left) that learns latent features $z$ without such distinction. 

Specifically, we propose a variational autoencoder with two latent feature spaces $z_r$ and $z_v$ respectively for robust and vulnerable features. We regularize $z_r$ and $z_v$ not to estimate distributions of adversarial and benign inputs respectively, in addition to an original objective of the variational autoencoder, as a mini-max game.
As the training converges, $z_r$ gets close to a distribution of robust features, and $z_v$ represents a distribution of vulnerable features. Then the $z_v$ could be used to detect adversarial inputs, since they will form a distinct distribution in $z_v$ that separates them from benign inputs. 

\subsection{Mini-max game}
A key idea of our proposed training can be represented as two-player mini-max game between a defender and an adversary on the two probability distributions of the robust features $P^c_r$ and the vulnerable features $P^c_v$ of each label $c$. 
The defender seeks to detect an adversarial attack on an input $x$ by checking if $P^c_v(x)$ is higher than a threshold, and the adversary aims to maliciously perturb the input $x$ while compromising the detection.
$$\min\limits_{\theta^{c}}\max\limits_{\delta \in S} logP^{c}_{r}(x + \delta) + log(1 - P^{c}_{v}(x + \delta))$$
In the beginning, $P^c_r$ is initialized to reflect the distribution of all the robust and vulnerable features,
but $P^c_v$ does not represent a feature.
Each player alternately plays the game.
In the adversary's turn, the adversary perturbs $x$ with $\delta \in S$, in order to maximize $P^c_r$ but minimize $P^c_v$ to compromise the defender. 
In the next turn, the defender controls the model parameter $\theta^{c}$ to perform the opposite, aiming to collect all the vulnerable features that
have been exploited by the adversary and segregate them into the distribution of $P^c_v$.
If the defender successfully detects all the vulnerable features, then the defender wins the game. Otherwise, the adversary wins. However, since the set of vulnerable features is finite with the finite size of $\theta^c$, the defender with a proper detection strategy will eventually win the game, after a sufficient number of turns.




\subsection{Network architecture}
\begin{figure}
  \centering
  \includegraphics[scale=0.51]{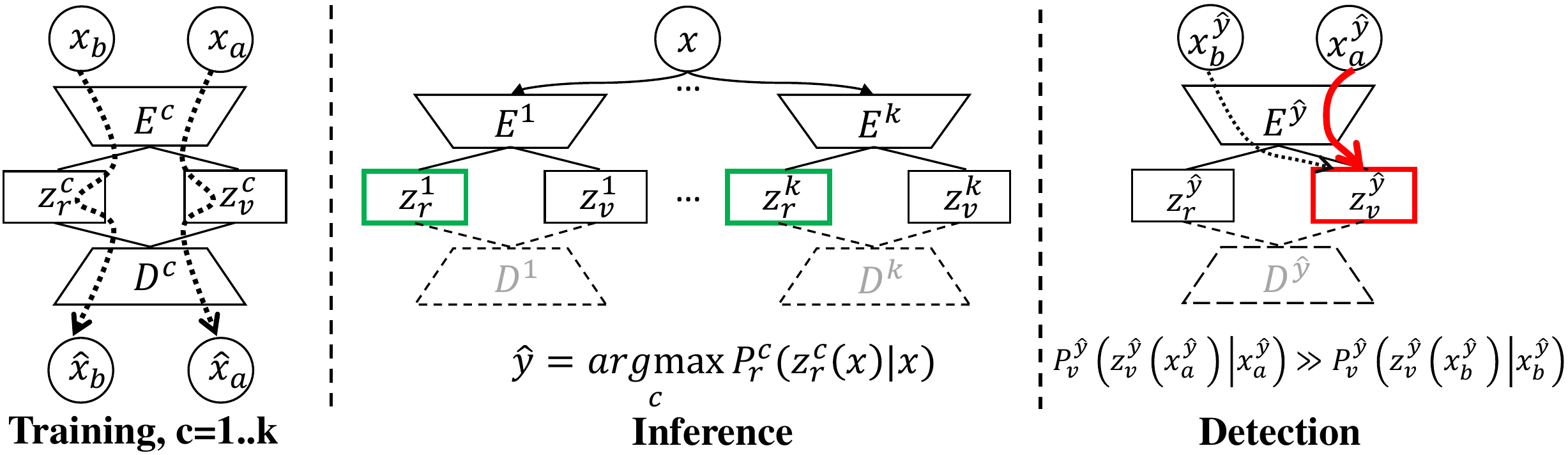}
\vspace{-0.1\baselineskip}
  \caption{Proposed network architecture for the disentangling learning}
\vspace{-1.8\baselineskip}
  \label{fig:network-architecture}
\end{figure}

To embed the proposed mini-max game in a training process,
we suggest a network architecture described in~\autoref{fig:network-architecture}.
For each label $c$, we have a variational autoencoder VAE$^{c}$\cite{vae}
that consists of an encoder $E^c(x)$ and  a decoder $D^c(x)$.
Instead of one type of latent variables, $E^c(x)$
samples two types of latent variables $z^c_{r}$ and $z^c_{v}$ for each of benign and adversarial inputs of the label $c$.
We denote $E^{c}(x)=E^{c}_r(x)\cdot E^{c}_v(x) = z^{c}_{r}(x) \cdot z^{c}_{v}(x)$ for
simplicity.
$D^c(z)$ generates $\hat{x}$ for any given $z$, where $z$ is either $z^c_r(x)$ or $z^c_v(x)$ depending on whether $x$ is $x_b$ or $x_a$.
The information is given in the training process as a flag, and $D^c(x)$ can selectively take
$z^c_{r}(x_b)$ or $z^c_{v}(x_a)$ based on the flag.

For classification, we integrate the VAE$^c$s into a classifier $F(x)$. Given an input $x$, $F(x)$ estimates probabilities of $x$ on
$P^c_r$ of each VAE$^c$, and returns an index of the highest probability
as the predicted label $\hat{y}$.
This process can be denoted as $\hat{y}=F(x)=arg\max\limits_{c}P^{c}_{r}(z^{c}_{r}(x)|x)$.
For the detection of an adversarial input $x_a$, we utilize the probability
of $x_a$ on $P^{\hat{y}}_v$ relatively compared to the values of $X^{\hat{y}}_b$,
given the predicted label $\hat{y}=F(x_a)$. Specifically, we detect if
$P_v^{\hat{y}}(z_v^{\hat{y}}(x)|x)\gg P_v^{\hat{y}}(z_v^{\hat{y}}(x_i)|x_i), \mathrm{where}\,x_i \in X_b^{\hat{y}}$.

\subsection{Training the network}
We train each VAE$^c$ with the loss $l^c$.
We design $l^c$ based on an evidence lower bound of $logP^c_r(x_b) + logP^c_v(x_a)$
with two regularization terms, which penalize errors in variational inference.  
Specifically, when erroneous estimates happen, that is, when $E^c_r$ assigns $x_a$ with high probability or $E^c_v$ assigns $x_b$ with high probability,
VAE$^c$ is penalized and encouraged to distinguish between the robust features of $x_b$ and the vulnerable features in $x_a$. 
We provide the detailed derivation of $l^c$ in Section~\ref{appendix:loss} of the \textbf{supplementary file}.
The final form of $l^{c}$ is as follows, where $\mu_{i}(z)$ selects the $i$-th mean element of $z$,
and $\sigma_i(z)$ selects the $i$-th standard deviation element of $z$ from reparameterization.
$$
\begin{aligned}
    l^c & = ||x_b - D^c(E^c_r(x_b))||^2+ ||x_a - D^c(E^c_v(x_a))||^2 \mathrm{\;\;\;\;\;\;\;\;\;\;\;\;\;\;\;\;\;\;\;\;\;\;\;\;\;\;\;(Reconstruction\;loss)}\\
            &-\frac{1}{2}\alpha\sum^{|E^c_r(x_b)|}_i (1 + log(\sigma^2_{i}(E^c_r(x_b))) - \mu^2_i(E^c_r(x_b)) - \sigma^2_i(E^c_r(x_b))) \mathrm{\;\;(KL\;divergence\;for\;} z_r^c)\\
            & - \frac{1}{2}\alpha\sum^{|E^c_v(x_a)|}_i (1 + log(\sigma^2_{i}(E^c_v(x_a))) - \mu^2_i(E^c_v(x_a)) - \sigma^2_i(E^c_v(x_a))) \mathrm{\;(KL\;divergence\;for\;} z_v^c)\\
            &-\beta\{log(1 - \mathcal{N}(E^c_r(x_a)|0,I)) + log(1 - \mathcal{N}(E^c_v(x_b)|0,I))\} \mathrm{\;\;\;\;\;\;\;\;\;\;\;(} z^c \mathrm{\;error\; penalty)} \\
\end{aligned}
$$
We choose the pixel-wise mean squared error (MSE) for the first two terms as reconstruction errors,
and the standard normal distribution $\mathcal{N}(0,I)$ as priors for $P^c_r$ and $P^c_v$.
We also introduce constants $\alpha$ and $\beta$, respectively, for the KL divergence terms and
the loss terms of variational inference for a practical purpose. 

Before we train VAE$^{c}$ with $l^c$  
we should incorporate the knowledge of our defense mechanism
to the existing attacks, considering the whitebox attack model.
Given an attack loss $l_{attack}$ of an arbitrary attack $A$,
we linearly combine a term
$l_{pass} = max(0,\;P_v^{\hat{y}}(z_v^{\hat{y}}(x_a)|x_a) - E_{x^{i} \in X^{\hat{y}}_{b}}[P_v^{\hat{y}}(z^{\hat{y}}_v(x^{i})|x^{i})])$ to bypass our detector with a coefficient $\gamma$.
As a result we get $l_{adapt\_attack} = l_{attack} \pm \gamma l_{pass}$.
We also introduce a binary search to find the proper $\gamma$ in a similar way to the CW-L2 attack. We modify all the attacks
in this paper. We denote an attack with "\_W", if the attack is
modified to work on the whitebox model. The sign of the $\gamma L_v$ term depends
on whether the attack is based on gradient descent (+), or ascent (-).

\setlength{\columnsep}{0pt}%
\begin{wrapfigure}{R}{0.5\textwidth}
\vspace{-1.5\baselineskip}
   \begin{minipage}{0.5\textwidth}
 \resizebox{1.0\textwidth}{!}{
\begin{algorithm}[H]
    \begin{algorithmic}
    \STATE $X^c_v \gets \phi:$ Set of adversarial inputs to label $c$
    \STATE $A\_W:$ An adapted whitebox attack
    \WHILE{$A\_W\;successes\;to\:bypass$}
        \STATE{$X^{c}_v \gets X^c_{v} \cup A\_W(X_b - X_b^c, \mathrm{VAE}^c) $}
        \STATE{\textbf{TRAIN} $\mathrm{VAE}^c$ to distinguish $X_b^c$ and $X^c_v$} maximizing $l^c$ with $\theta^c$
    \ENDWHILE
\end{algorithmic}
  \caption{Training process of VAE$^c$}
\vspace{-0.2\baselineskip}
    \label{algo:training-vaec}
\end{algorithm}
}                                        
   \end{minipage}
\end{wrapfigure}
Algorithm~\ref{algo:training-vaec} describes our training algorithm.
It is an iterative process, where at each iteration $A\_W$ attacks VAE$^c$ to exploit vulnerable features in $P^c_r$, and the VAE$^c$ corrects the distribution $P^c_r$ and $P^c_v$ with $\theta^c$ to identify vulnerable features found in each iteration. We should only attack $X_b - X^c_b$, in order to prevent the inclusion of robust features in $P^c_v$.
The training ends when $A\_W$ can not find adversarial inputs that could bypass the detector.


\section{Experiment}

\setlength{\columnsep}{5.0pt}%
\begin{wraptable}{R}{0.45\textwidth}
\vspace{-1.0\baselineskip}
   \begin{minipage}{0.45\textwidth}
       \caption{$(X_b, Y_b)$ accuracy of models} 
    \resizebox{1.0\textwidth}{!}{
%

%
%

\begin{tabular} {c c c c}
\toprule
     & MIM\_W & CW-L2\_W& No defense\\
\midrule
    MNIST       &0.97& 0.98& 0.99\\
\midrule
    Fashion     &0.98& 0.98& 0.99\\
    MNIST5      \\
\midrule
    Cat \& Dog & 0.96& 0.96 & 0.99\\

\bottomrule 

\end{tabular}

        \label{table:accuracy}
    }
        \centering
   \end{minipage}
\vspace{-1.0\baselineskip}
\end{wraptable}

We evaluate our defense mechanism under 
both blackbox and whitebox attacks, trained with MIM\_W and CW-L2\_W.
In the blackbox setting, we evaluate how precisely 
our detector filters out adversarial inputs by measuring AUC scores.
In the whitebox setting, we first quantitatively evaluate attack success ratio and qualitatively analyze whether 
successful adversarial inputs induce semantic changes.

\textbf{Baseline.}
We choose Gong et al.~\cite{gong-detect} as a baseline for comparison,
which also leverages adversarial inputs to train an auxiliary classifier
for detecting adversarial inputs.
Note that Gong et al. does not incorporate adaptive attacks in their approach,
although we denote it with the same notation (e.g., MNIST, A=PGD\_W).

\textbf{Attacks.}
Our attacks are based on the publicly available implementations~\cite{mnistchallenge,cleverhans},
and the whitebox attacks are adapted to bypass our detection mechanism.
All adversarial inputs are generated from a separate test dataset, in an untargeted way.

\textbf{Datasets.}
We evaluate our detector on the MNIST~\cite{mnist}, Fashion MNIST5~\cite{fashion-mnist}, and Cat \& Dog~\cite{cat-and-dog} datasets.
In the case of the Cat \& Dog dataset, we collect total 2028 frontal faces of
cats and dogs, and resize it into 64 x 64 x 1 with a single channel.

We first show that our defense methods achieve a level of accuracy similar to those without defense mechanisms (see \autoref{table:accuracy}). Additional information including training parameters and details of the attacks are described in Section~\ref{appendix:expinfo} of the \textbf{supplementary file}.


\subsection{Blackbox substitute model attack}
In the blackbox setting, the adversary has no information regarding 
our defense mechanism, but we assume that the adversary has the same datasets as a defender.
The adversary builds its own standard substitute classifier $F_s$,
and generates a group of adversarial inputs $x_a$ with an attack $A$ to fool $F_s$.
After that, the adversary attacks $F$ with $x_a$, and the defender detects $x_a$ based on the values of $P^{\hat{y}}_v(z^{\hat{y}}_v(x_a)|x_a)$ where $\hat{y} = F(x_a)$.
The blackbox substitute model attacks are possible exploiting the transferability~\cite{fgsm,delving,ilyas,narod,zoo} of $x_a$ over classifiers trained with similar datasets.

\autoref{table:blackbox-attack} shows AUC scores for detection results, where each cell compares ours (left) with the baseline (right) for each individual attack while 
our defense approaches achieve 0.98 on average and perform better than the baseline by up to 0.33.
We attribute the success of our model to its ability to disentangle the distribution of the vulnerable features of $x_a$ into $z_v$ from the distribution of whole features. To see if the features are actually disentangled, we visualize $z^c$ with t-SNE~\cite{t-SNE} in Figure 5. It shows clear separation between $z^c_r(x^c_b)$ and $z^c_v(x^c_a)$ as we expected. We conclude that the transferability between the models is reduced by disentangling the vulnerable features which the adversary might exploit for $F_s$ found in benign datasets.
\begin{table}
    \parbox{.55\linewidth}{
    	\vspace{-1.1\baselineskip}
        \caption{\small{AUC scores of blackbox attack detection. Our approach (left) remains more generalized over various attacks compared to the baseline (right)}}
        \centering
        \resizebox{0.55\textwidth}{!}{
%
%
%
%
%
%

\begin{tabular} {c c c c c}

\toprule
    &\multicolumn{2}{c}{\textbf{MNIST}} & \multicolumn{2}{c}{\textbf{Fashion MNIST5}} \\
    \cmidrule(r){2-3} \cmidrule(r){4-5} 
    &    MIM\_W& CW-L2\_W  &    MIM\_W& CW-L2\_W               \\
\midrule
    FGSM &    0.99 / 0.99&  0.98 / 0.99        &0.98 / 0.99  & \textbf{0.99} / 0.97                  \\
    PGD &\textbf{0.99} / 0.96&      0.99 / 0.99         &\textbf{0.97} / 0.79 & \textbf{0.99} / 0.96    \\
    MIM &\textbf{0.99} / 0.90&   \textbf{0.99} / 0.98            &\textbf{0.98} / 0.92   & \textbf{0.99} / 0.97   \\
    CW &\textbf{0.97} / 0.64&   \textbf{0.97} / 0.96             &\textbf{0.96} / 0.64  &  \textbf{0.97} / 0.95 \\
\bottomrule 

\end{tabular}

        }
    \label{table:blackbox-attack}
    }
    \begin{minipage}{0.40\linewidth}
    \centering
        \centering
        \includegraphics[width=1.0\textwidth]{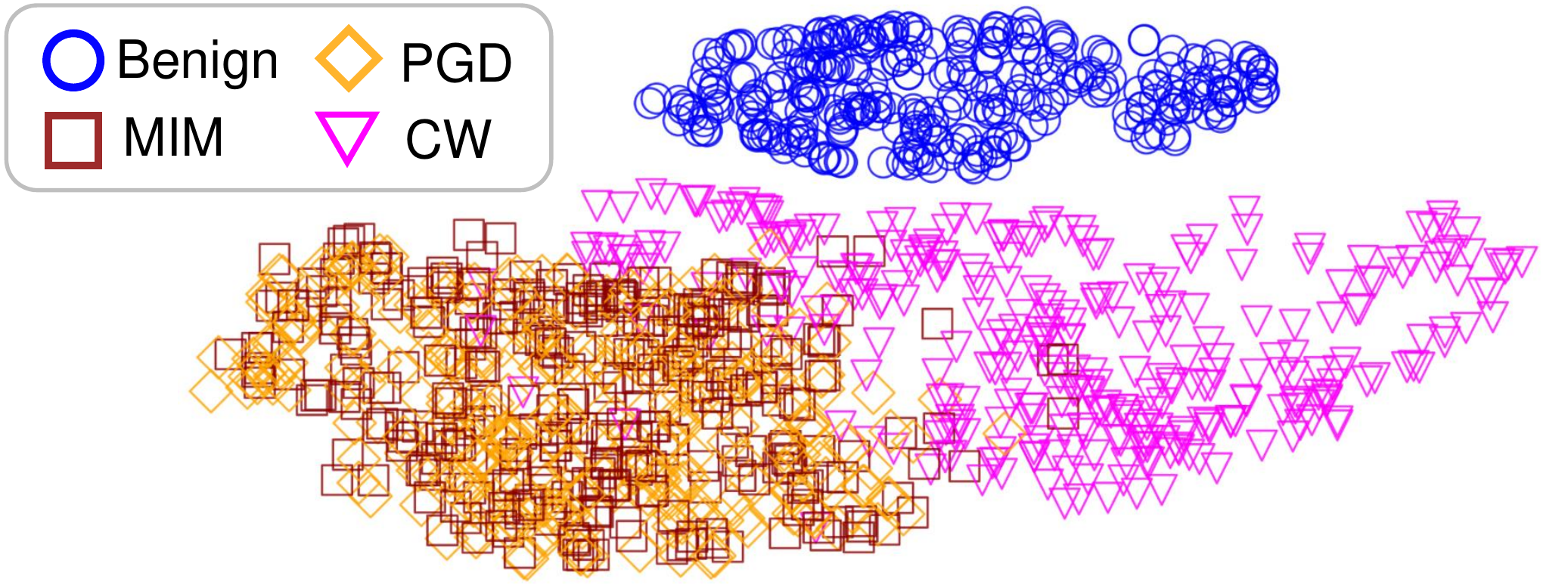}
        \captionof{figure}{\small{Disentangled latent distributions of robust and vulnerable features  (blue $x_b$ vs. others $x_a$), on MNIST, A=MIM\_W VAE$^{c=3}.$}}
        \label{fig:disentangled}
    \end{minipage}
\vspace{-1.2\baselineskip}
\end{table}

\subsection{Whitebox attack}
Whitebox attacks are difficult to defend because the adversary has exactly the same knowledge as the defender, which could be exploited in order to fool the defender. 
For clear analysis, we define a set of success conditions $C$ of the adversary when an inference label of $x_a$ is $\hat{y}=F(x_a)$, as follows:
\begin{enumerate}[label=$C\arabic*$]
    \item Low probability on vulnerable features to bypass the detector: $P^{\hat{y}}_v(x_a) < E[P^{\hat{y}}_v(x^{\hat{y}}_b)]$. 
    \item High probability on robust features  to convince the defender: $P^{\hat{y}}_r(x_a) > E[P^{\hat{y}}_r(x^{\hat{y}}_b)]$.
\item Semantic meaning of the original input should be retained. 
\end{enumerate}

\begin{figure}
  \centering
  \includegraphics[scale=0.265]{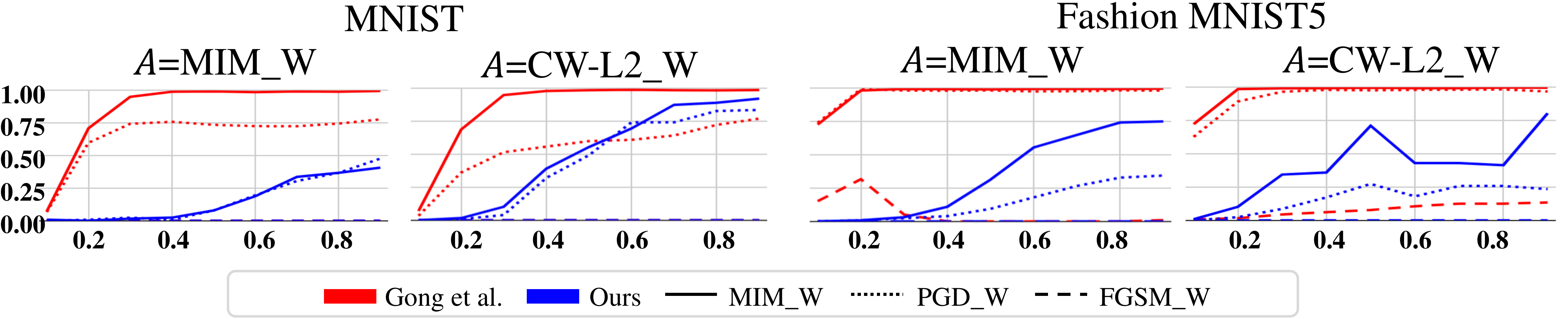}
\vspace{-0.4\baselineskip}
    \caption{\small{Success ratio of whitebox attacks along the $L_\infty$ distortion (according to C1 \& C2)}}
\vspace{-0.5\baselineskip}
  \label{fig:whitebox-mnists-eps}
\end{figure}

\setlength{\columnsep}{0pt}%
\begin{wraptable}{R}{0.45\textwidth}
\vspace{-1.5\baselineskip}
   \begin{minipage}{0.45\textwidth}
        \caption{Result of CW-L2\_W attack}
        \resizebox{0.9\textwidth}{!}{
%
%

\begin{tabular} {l c c c c}

\toprule
               &\multicolumn{2}{c}{\textbf{MNIST}} & \multicolumn{2}{c}{\textbf{Fashion MNIST5}} \\
    \cmidrule(r){2-3} \cmidrule(r){4-5}
          A    & MIM\_W & CW-L2\_W & MIM\_W &CW-L2\_W  \\
\midrule
    Ratio      &0.32   & 0.28   &0.18    &0.19                \\
    Mean $L_2$ &45.08  &48.25   &43.34   &37.66      \\
\bottomrule 

\end{tabular}

        }
        \label{table:cw-attack}
        \centering
   \end{minipage}
\vspace{-1.0\baselineskip}
\end{wraptable}
For $C1$ and $C2$, \autoref{fig:whitebox-mnists-eps} plots
attack success ratios along the $L_\infty$ distortion on MNIST and Fashion MNIST5.
As $L_\infty$ increases, the success ratio also increases except FGSM\_W.
Our defense shows gradual slope compared to the baseline.
\autoref{table:cw-attack} shows the whitebox attack result of CW-L2\_W with minimized distortions in a binary search. CW-L2\_W achieves average success ratio of 0.30 and 0.19 with $L_2$ distortion\footnote{The L$_2$ distortion is calculated as $||\delta||_2/\sqrt{d}$ in [0, 255] input range.} of 46.66 and 40.5 for MNIST and Fashion MNIST5, respectively.

\begin{figure}[t]
  \centering
  \includegraphics[scale=0.42]{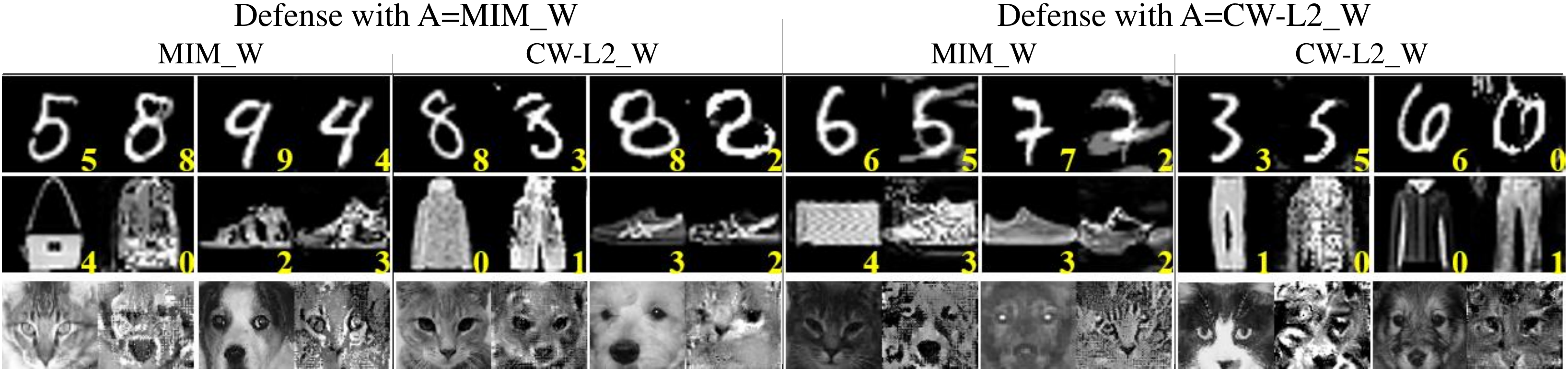}
\vspace{-1.4\baselineskip}
  \caption{\small{Visual results of whitebox attacks on our defense}}
\vspace{-1.5\baselineskip}
  \label{fig:whitebox-mnists}
\end{figure}

\begin{figure}

\vspace{-1.2\baselineskip}
  \centering
  \includegraphics[scale=0.43]{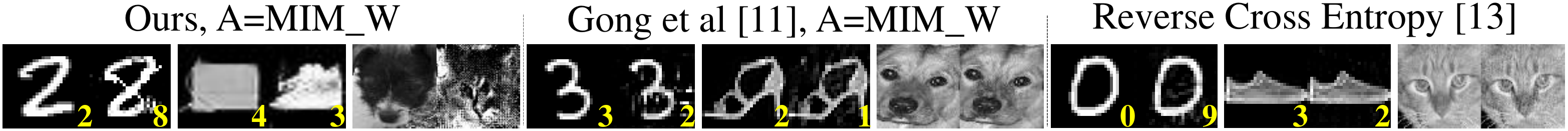}
\vspace{-0.2\baselineskip}
    \caption[Caption for LOF]{\small{Semantic comparison to baselines\cite{gong-detect,rce} under the CW-L2\_W attack;
    	the attack yields larger semantic changes on our approach compared to the 
baselines}\protect\footnotemark}
\vspace{-1.0\baselineskip}
  \label{fig:gongetal-semantics-cmp}
\end{figure}

Regarding $C3$, \autoref{fig:whitebox-mnists} compares the visual differences between each pair of a benign image (left) and an adversarial image (right). The predicted label for each image is shown in yellow, in the bottom right corner. We choose $L_\infty=0.5$ as a reference distortion value for the MIM\_W attacks.
We can clearly observe the semantic changes on the adversarial images.
We additionally evaluate our defense mechanism on a Cat \& Dog dataset.
It also shows the semantics changed between the labels ($L_{\infty}=0.3$).
Some lines are appeared or disappeared in MNIST rather than
noisy dots, and sneakers turn into sandals with similar styles such as overall shape or pattern.
In the case of Cat \& Dog dataset, features of dogs are appearing
in the adversarial inputs generated from cats as big noses and long spouts,
while the brightness of the fur, or angles of faces seem to be preserved.
\autoref{fig:gongetal-semantics-cmp} show adversarial perturbations from the attacks result in clear semantic changes with our approach compared to other baselines. We provide more results obtained with various $L_{\infty}$ including the PGD\_W attack in Section~\ref{appendix:expfigs} of the \textbf{supplementary file}.
From the result we conclude our approach successfully distinguishes vulnerable features from whole features compared to the baseline.
Furthermore, considering the semantic changes of adversarial inputs,
we conclude that the robust features estimated in $z_r$ are well-aligned with human perception.



\footnotetext{We found and corrected implementation errors in the robustness evaluation of the reverse cross entropy~\cite{rce}, and we could bypass the detector in the adaptive whitebox attack. We confirmed it with an author of the paper.}


%
%
%
%


\section{Conclusion}
In this work, we hypothesize about the latent feature space for adversarial inputs of a label and conjectured that feature space entanglement of vulnerable and robust features is the main reason of adversarial vulnerability of neural networks, and proposed to \emph{learn} a space that disentangles the latent distributions of the vulnerable and robust features. Specifically, we trained a set of variatonal autoencoders for each label with two latent spaces, and trained them using a two-player mini-max game, which results in learning disentangled representations for robust and vulnerable features. We show that our approach successfully identifies the vulnerable features and also identifies sufficiently robust features in the whitebox attack scenario. However, we cannot guarantee that our approach is a panacea, and further research is required for the discovery of new attacks. We hope our work stimulates research toward more reliable and explainable machine learning.

\clearpage

\bibliographystyle{utphys}
\bibliography{p}
\clearpage
\appendix
\section{Appendix}
\subsection{Additional information about the motivational experiment}

We conducted the motivational experiments described in the Section 3.1 to demonstrate that 
the classifier $F_v$ trained with a dataset ($X_v$,$Y_v$) only containing vulnerable features can achieve high accuracy on 
a benign dataset ($X_b$, $Y_b$). Table~\ref*{table:moti-model-params} and Table~\ref{table:moti-attack-params} describe the model parameters and the attack parameters used in the experiments. Algorithm~\ref{algo:intro:vfr-training} details the process of creating ($X_v$,$Y_v$) with the discriminator $D$.

We use abbreviated notations in \autoref{table:moti-model-params}, and \autoref{table:moti-attack-params}.
The "c(x,y)" is a convolutional layer with ReLU activation. The x is size of a kernel, 
and the y is the number of kernels.
"mp(x)" is a max pooling layer whose pooling size is x by x.
The "d(x)" is a dense layer where x is the number of neurons.
The "sm(x)" is a softmax layer with output dimension x.
For the attack parameter, "e" is $L_\infty$ epsilon and i is iteration of attacks.
"ss" is a step size of perturbations.
"df" is a decay factor for the MIM attack. 
"lr", "cf", "ic", "bs" are learning rate, confidence, initial coefficient for the miss-classification loss, and the number of binary search steps.

For the fashion MNIST5, we intentionally choose the subset of Fashion MNIST such as coat (0), trouser (1), sandal (2), sneaker (3), bag (4) 
for decreasing effect of inter-label robust features. For example, sneaker and ankle boot, coat and pull over are quite similar. By doing so, we could extract vulnerable features only for each label and get an accurate result. 

We implement the discriminator $D$ as a separate model with one dimension of sigmoid output.
The $D$ learns to distinguish benign inputs as 0, and adversarial inputs as 1.
As a conequence, we can interpret the output value of $D$ as a probability where an input $x$ would be an adverarial input.
In terms of the bypassing $D$ for attacks, we linearly incorporate the output value of $D$
in objective loss functions of the attacks.
As the attacks minimizing the output probability of $D$, the attacks generate new $x_a$ with new vulnerable features.

  %
  %

\begin{table}[h]
  \scriptsize
  \centering
    \caption{Model parameters in the motivational experiment}
\begin{tabular} {l c }

\toprule
 \textbf{Models}                    &\textbf{Parameters}      \\
\midrule
    MNIST, $F_p$           &c(2,20) mp(2) c(2,50) mp(2) d(500) sm(10)    \\
    MNIST, $F_v$          & c(5,20) mp(2) c(5,50) mp(2) d(256) sm(10) \\
    \midrule
    Fashion MNIST5, $F_p$   &c(5,20) mp(2) c(5,50) mp(2) d(500) sm(10)  \\
    Fashion MNIST5, $F_v$ &c(5,20) mp(2) c(5,50) mp(2) d(256) sm(10) \\
\bottomrule 

\end{tabular}

    \label{table:moti-model-params}
\end{table}

\begin{table}[h]
  \scriptsize
  \centering
    \caption{Attack parameters in the motivational experiment}
\begin{tabular} {l c c c c}

\toprule
                 & \multicolumn{4}{c}{\textbf{Blackbox substitute model attacks}}                               \\
                                            \cmidrule(r){2-5}                                                                                                                
                 & \textbf{FGSM}     & \textbf{PGD}          & \textbf{MIM}               & \textbf{CW-L2}      \\
\midrule
    MNIST        &e:0.3              &e:0.3, i:90,           &e:0.3, i:640,               &i:160, lr:0.1        \\
                 &                   &ss:0.01                &ss:0.01, df:0.3             &cf:3, ic:10, bs:1    \\
\midrule
    Fashion      &e:0.3              &e:0.4, i:90            &e:0.3 , i:320               &i:160, lr:0.1        \\
    MNIST5       &                   &ss:0.01                &ss:0.001, df:0.3             &cf:3, ic:10, bs:1    \\
\bottomrule 
\end{tabular}

%


    \label{table:moti-attack-params}
\end{table}

\label{appendix:expvfr}
\begin{algorithm}[h]
\begin{algorithmic}
    \STATE ($X_b$, $Y_b$): Given benign dataset 
    \STATE ($X_v$, $Y_v$) $\gets (\phi, \phi)$: Empty dataset for vulnerable features
    \STATE $F, F':$ Pre-trained and initialized model respectively with same input and output dimensions
    \STATE $A:$ Arbitrary attack
    \STATE $D:$ Discriminator between benign (0) and adversarial (1) inputs 
    
    \WHILE{$i < limit $}
        \STATE{$X^{i}_v,~Y^{i}_v \gets A(X_b, Y_b, F, D) $}
        \STATE{$(X_{v}, Y_{v}) \gets (X_{v} \cup X^{i}_{v}, Y_{v} \cup Y^{i}_{v}$)} 
        \STATE{\textbf{TRAIN} $D$ to distinguish $X_b$ and $X_v$}
    \ENDWHILE
    \STATE{\textbf{TRAIN} $F'$ with $X_v$ and $Y_v$}
    \STATE{\textbf{PRINT} accuracy of $F'$ on $X_b$ and $Y_b$}
     
\end{algorithmic}
  \caption{Training process only with the vulnerable features $X_v$}
    \label{algo:intro:vfr-training}
\end{algorithm}

\subsection{Training loss derivation}
\label{appendix:loss}
To approximate and distinguish the latent variable distributions 
of $z^c_r$ and $z^c_v$, we maximize $ELBO(L^{c})$ for each VAE$^{c}$, where

\begin{equation}
    \label{eq:lc}
        L^{c} = L^{c}_{E} + L^{c}_{I}
\end{equation}
$L^{c}$ consists of two terms. The first one is an evidence term $L^c_{E}$
indicating the probability of occurence of $x_b$ and $x_a$, where

\begin{equation}
\begin{aligned}
        L^{c}_{E} = logP^c_r(x_b) + logP^c_v(x_a)
\end{aligned}
\end{equation}

The second one is a loss term of variational inference which penalizes
in the case of wrong variation inference to each distribution $P^c_r$ and $P^c_v$.
It can be expanded to incorporate latent variables $z^c_r$ and $z^c_v$ as follows.
\begin{equation}
\begin{aligned}
    L^{c}_{I} &= log(1 - P^c_r(x_a)) + log(1 - P^c_v(x_b)) \\
                  &= log(1 - \sum_{z^c_r(x_a)}P^c_r(x_a)P^c_r(z^c_r(x_a)|x_a)) + log(1 - \sum_{z^c_v(x_b)}P^c_v(x_b)P^c_v(z^c_v(x_b)|x_b)) \\
                  &= log(1 - E[P^c_r(z^c_r(x_a)|x_a)]) + log(1 - E[P^c_v(z^c_v(x_b)|x_b)]) \\
\end{aligned}
\end{equation}

Plugging typical ELBO expansion~\cite{vae} of the $L^c_{E}$ term, and the $L^c_{I}$ term into the equation \ref{eq:lc},
we get following $ELBO(L^c)$.
\begin{equation}
\begin{aligned}
    L^{c} \geq&E[logP^c_r(x_{b}|z^{c}_r(x_{b}))]+E[logP^c_v(x_{a}|z^{c}_v(x_{a}))] \\
            &- KL[q^c_r(z^{c}_r(x_{b})|x_{b})||P^c_r(z^{c}_r(x_{b}))] - KL[q^c_v(z^{c}_v(x_{a})|x_{a})||P^c_v(z^{c}_v(x_{a}))] \\
            &+log(1 - E[P^{c}_r(z^{c}_r(x_a)|x_a)]) + log(1 - E[P^{c}_v(z^{c}_v(x_b)|x_b)]) \\
            &=ELBO(L^c)
\end{aligned}
\end{equation}

\begin{equation}
\begin{aligned}
    l^c = &||x_b - D^c(E^c_r(x_b))||^2+ ||x_a - D^c(E^c_v(x_a))||^2\\
            &-\frac{1}{2}\alpha\sum^{|E^c_r(x_b)|}_i (1 + log(\sigma^2_{i}(E^c_r(x_b))) - \mu^2_i(E^c_r(x_b)) - \sigma^2_i(E^c_r(x_b)))\\
            & - \frac{1}{2}\alpha\sum^{|E^c_v(x_a)|}_i (1 + log(\sigma^2_{i}(E^c_v(x_a))) - \mu^2_i(E^c_v(x_a)) - \sigma^2_i(E^c_v(x_a))) \\
            &-\beta\{log(1 - \mathcal{N}(E^c_r(x_a)|0,I)) + log(1 - \mathcal{N}(E^c_v(x_b)|0,I))\} \\
\end{aligned}
\end{equation}
We choose the pixel-wise mean squared error (MSE) for the first two terms as reconstruction errors,
and the standard normal distribution $\mathcal{N}(0,I)$ as priors for $P^c_r$ and $P^c_v$.
We also introduce constants $\alpha$ and $\beta$, respectively, for the KL divergence terms and
the loss terms of variational inference for a practical purpose. 

\subsection{Training and attack parameters}
\label{appendix:expinfo}
We use abbreviated notations in~\autoref{table:model-params} as like in~\autoref{table:moti-model-params}
In additionto that the "z(x,y)" is a sampling layer for latent variables.
The x and y are dimensions of $z_r$ and $z_v$.
We use $\alpha=1$ and $\beta=100$ in all trainings.
For the label inference we choose the nearest mean classifier on $P^c_r$, because its linear property prevents
the vanishing gradients problem which makes attacks fail
but known to be penetrable.

\begin{table}[h]
  \scriptsize
  \centering
    \caption{Training parameters and accuracy of $X_b$ with/without our defense. (e:$L_\infty$ distortion, i:iterations, ss:step\_size, df:decay\_factor, bs:binary\_search\_steps, lr:learning\_rate, cf:confidence, ic:initial\_constant)}
\begin{tabular} {l c c}

\toprule
                     &\textbf{Model} & \textbf{Parameters}    \\
\midrule
    MNIST, $A$=MIM\_W            &$E^c$:3(c(4,16))z(8,8)  &e:0.5, i:3e3, ss:1e-3, df:0.9, bs:0  \\
    MNIST, $A$=CW-L2\_W          & $D^c$:d(24)d(49)3(ct(4,16))d(784) &i:1e3, lr:1e-3, cf:0, ic:1, bs:0      \\
    \midrule
    Fashion MNIST5, $A$=MIM\_W   &$E^c$:3(c(4,32))z(8,8)  &e:0.5, i:3e3, ss:1e-3, df:0.9, bs:0   \\
    Fashion MNIST5, $A$=CW-L2\_W &$D^c$:d(24)d(49)3(ct(4,32))d(784) &i:2e3, lr:1e-3, cf:0, ic:1, bs:0                                     \\
    \midrule
    Cat \& Dog, $A$=MIM\_W   &$E^c$:2(c(12,32)-bn-relu-mp(2))c(12,32)-bn-relu-z(64,64)  &e:0.2, i:12e2, ss:1e-3, df:0.9, bs:0   \\
    Cat \& Dog, $A$=CW-L2\_W &$D^c$:d(24)d(49)3(ct(4,64))d(4096)                         &i:3e3, lr:3e-3, cf:0, ic:1, bs:0                                     \\
\bottomrule 

\end{tabular}

    \label{table:model-params}
\end{table}

\begin{table}[h]
  \scriptsize
  \centering
  \caption{Attack parameters used in the experiments}
    \resizebox{\textwidth}{!}{
\begin{tabular} {l c c c c c c c c}

\toprule
                          & \multicolumn{4}{c}{\textbf{Blackbox substitute model attacks}}                                                   & \multicolumn{4}{c}{\textbf{Whitebox attacks}}                                \\
                                            \cmidrule(r){2-5}                                                             \cmidrule(r){6-9}                                                   
                          & \textbf{FGSM} & \textbf{PGD} & \textbf{MIM}  & \textbf{CW-L2}                                & \textbf{FGSM\_W} & \textbf{PGD\_W} & \textbf{MIM\_W}     & \textbf{CW-L2\_W} \\
\midrule
    MNIST        &e:0.3              &e:0.4, i:90,        &e:0.3, i:160,              &i:160, lr:0.1        &bs:3              &i:240, ss:0.01,             &i:1200,bss:1e-3,         &i:1.2e4, lr:0.01      \\
            &                    &ss:0.01             &ss:0.01, df:0.3            &cf:3, ic:10, bs:1    &                  &bs:3                            &df:0.9, bs:3                &cf:200, ic:10, bs:3    \\
\midrule
    Fashion    &e:0.3              &e:0.4, i:90            &e:0.3 , i:160               &i:160, lr:0.1                   &bs:3                 &i:240, ss:0.01,               &i:1200, ss:1e-3,                   &i:1.2e4, lr:0.01                 \\
    MNIST5       &                 &ss:0.01                &ss:0.01, df:0.3             &cf:3, ic:10, bs:1               &                     & bs:3      & df:0.9, bs:3          &cf:200, ic:10, bs:3                 \\
\bottomrule 
\end{tabular}

%


    }
    \label{table:mnists-attack-params}
\end{table}

\subsection{Additional figures about the whitebox attacks}
    \label{appendix:expfigs}

\begin{figure}[h]
  \centering
  \includegraphics[scale=0.46]{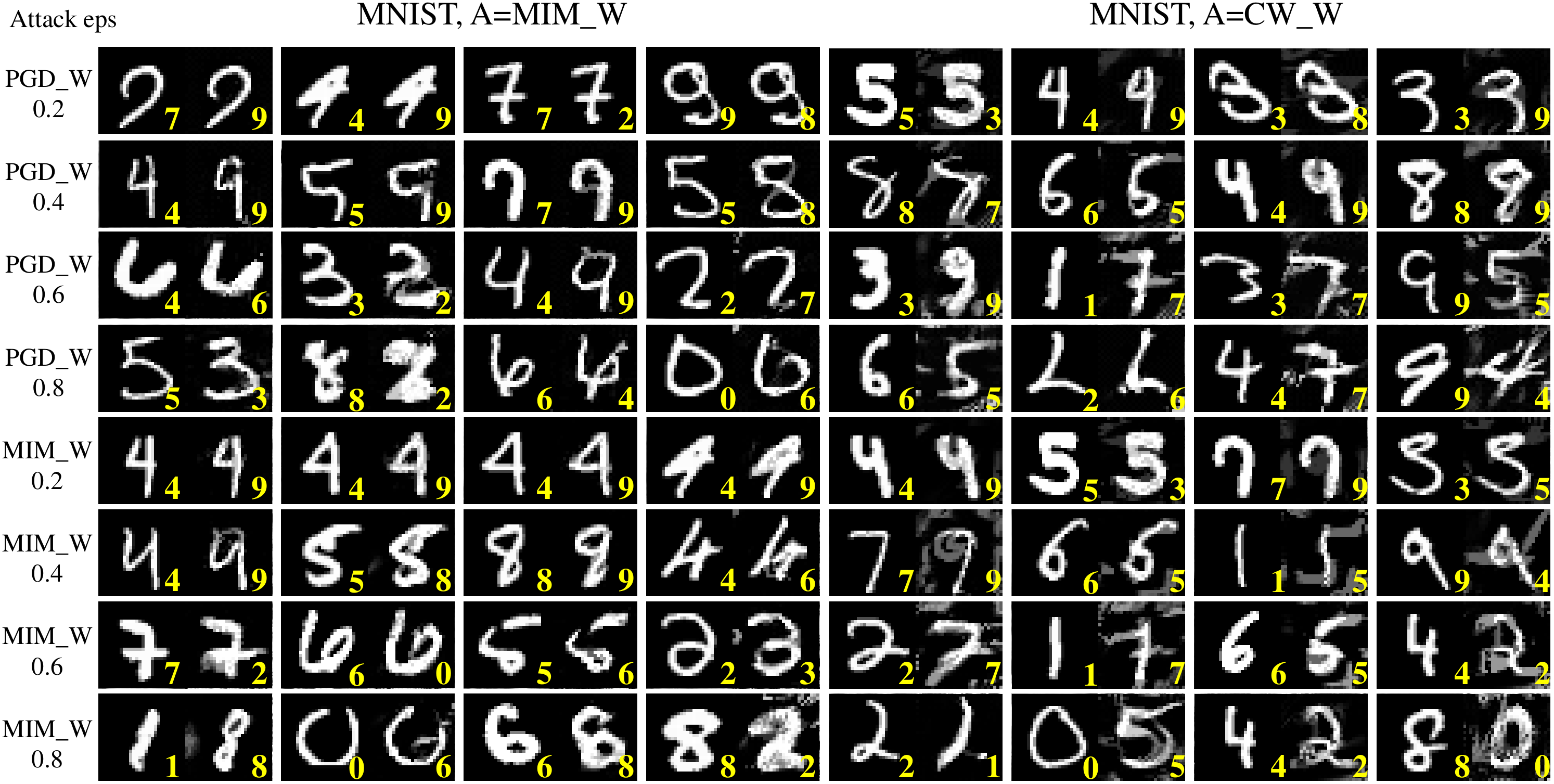}
  \caption{Whitebox attack changes semantics of inputs (MNIST)}
  \label{fig:appendix-whitebox-mnist-semantics}
\end{figure}

\begin{figure}[h]
  \centering
  \includegraphics[scale=0.46]{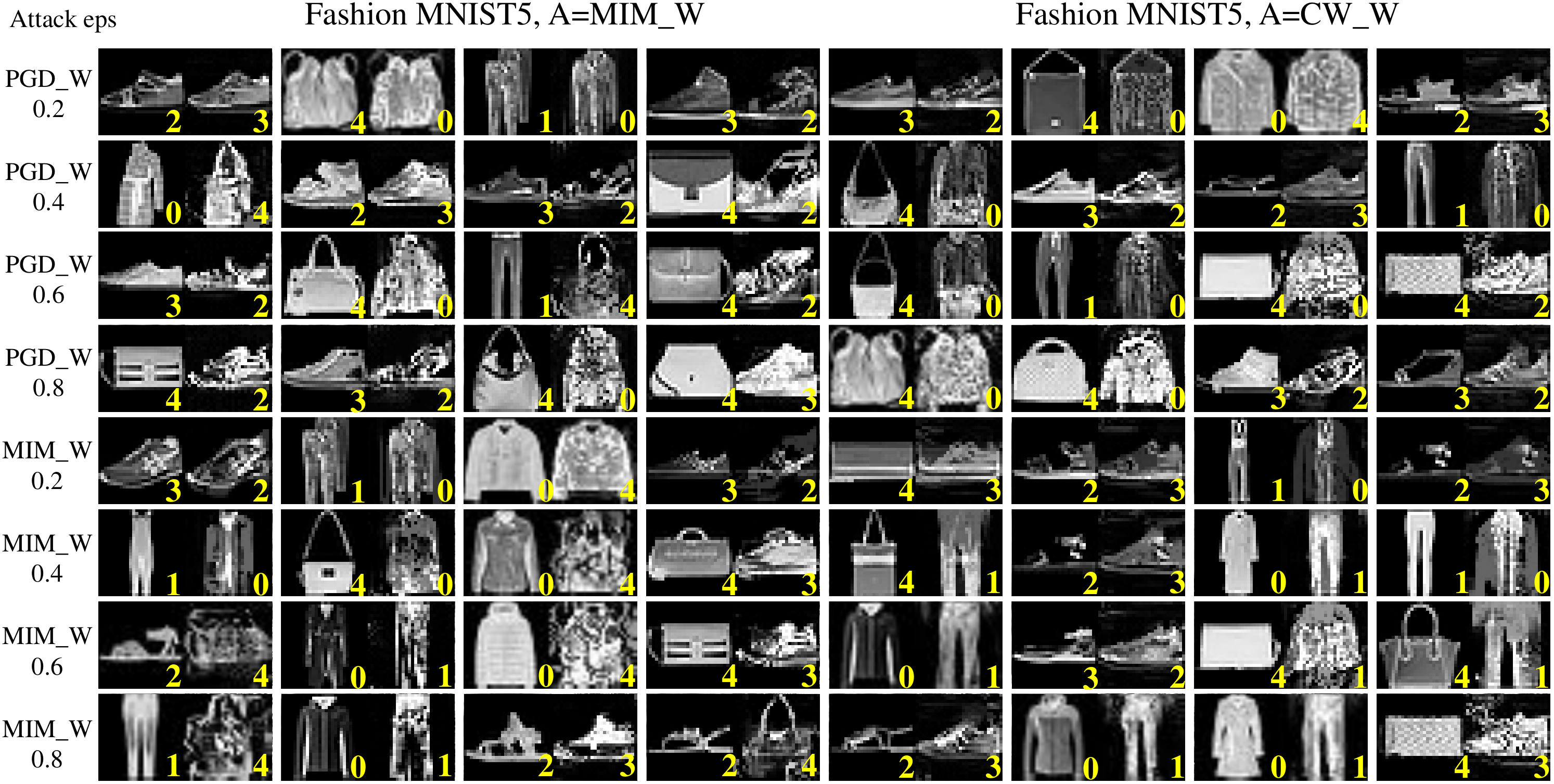}
  \caption{Whitebox attack changes semantics of inputs (Fashion MNIST)}
  \label{fig:appendix-whitebox-fashion-mnist-semantics}
\end{figure}

\begin{figure}[h]
  \centering
  \includegraphics[scale=0.46]{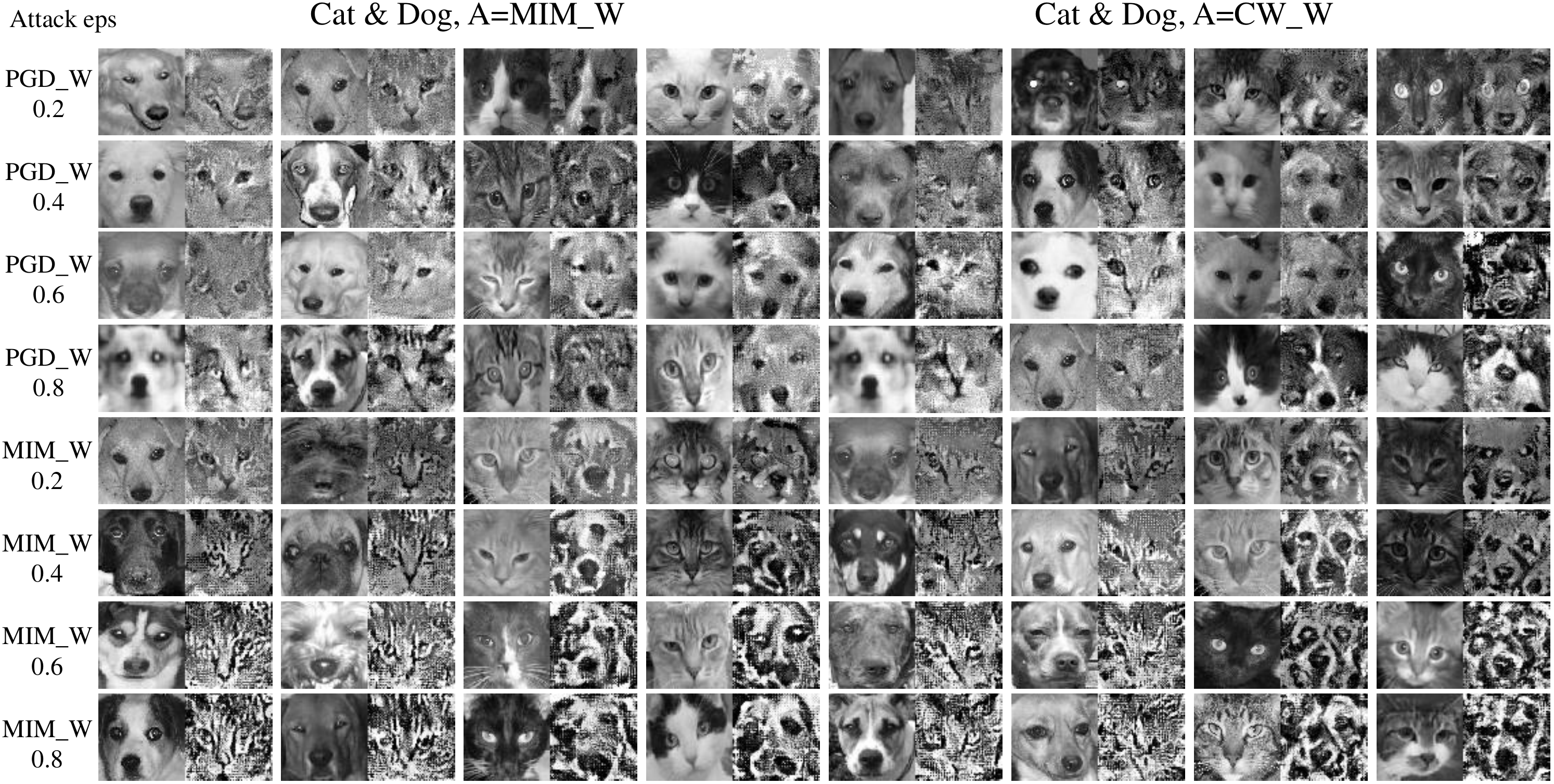}
  \caption{Whitebox attack changes semantics of inputs (Cat \& Dog)}
  \label{fig:appendix-whitebox-cat-and-dog-semantics}
\end{figure}

In this section, we qualitatively evaluate the performance of our proposed defense mechanism against whitebox attacks as a function of epsilon. Figures~\ref{fig:appendix-whitebox-mnist-semantics},~\ref{fig:appendix-whitebox-fashion-mnist-semantics}, and~\ref{fig:appendix-whitebox-cat-and-dog-semantics} show the results of the PGD\_W and MIM\_W attacks under a wide range of the epsilon from 0.2 to 0.8. Even when the value of the epsilon is small, there are many cases where one may make mistake.
As the epsilon becomes larger, semantic changes become more apparent. In the case of MNIST, attacks frequently occurred to 4, 7, 9, 3, and 5, which are of similar shapes. In Fashion MNIST5, attacks also frequently occurred to similar forms such as sandals and sneakers. The attack between the sandals and the sneakers shows that the original style is maintained to some extent, and a new image is created. In Cat \& Dog, when a cat image was attacked towards a dog, it was found that dog nose and long spout typically appeared. On the other hand, when dog images were attacked towards a cat, flat nose and Y-shaped mouth appeared.
Especially, when the epsilon is very large up to 0.8, the robust and the vulnerable features are well learned when we see that the semantically meaningful change is dominant and no perturbation like noise is added in the background.





[h]

\begin{figure}
  \centering
  \includegraphics[scale=0.46]{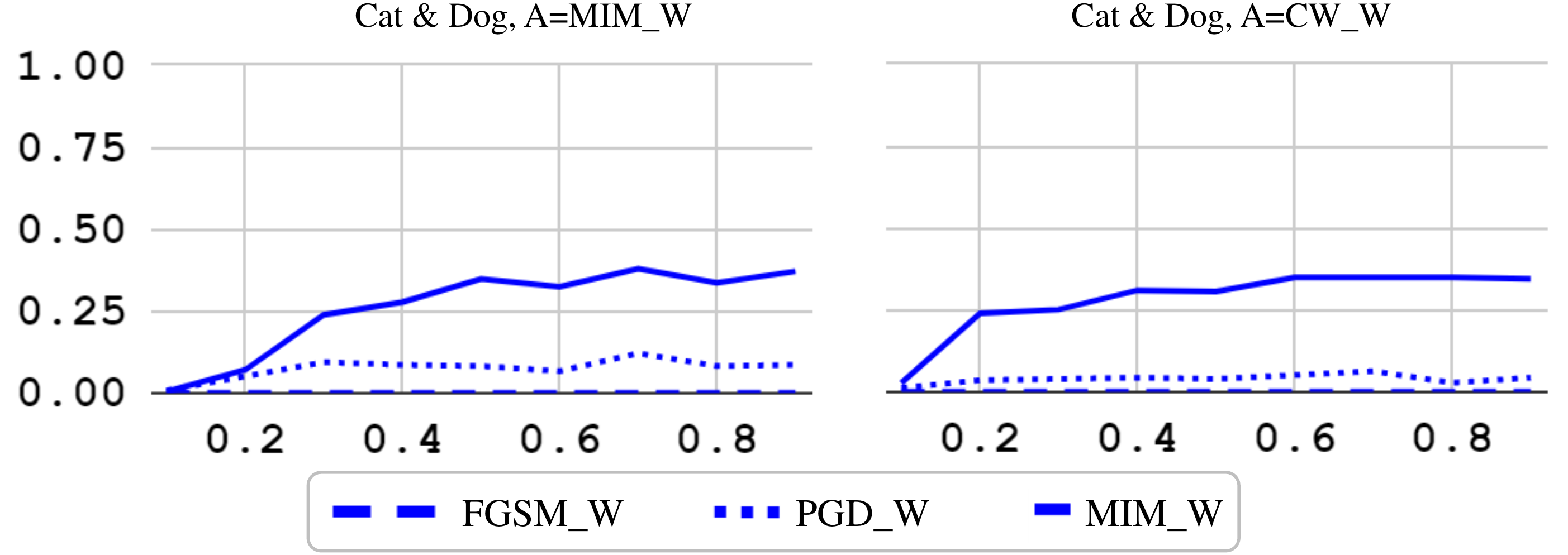}
    \caption{Whitebox attack success ratio along the $L_\infty$ distortion (Cat \& Dog).}
  \label{fig:appendix-cat-dog-eps}
\end{figure}

Figure~\ref{fig:appendix-cat-dog-eps} depicts the attack success ratio (according to C1 \& C2) on the Cag \& Dog dataset as a function of epsilon. Unlike MNIST and Fashion MNIST5, the success ratio is not so high. This can be interpreted to mean that the Cat \& Dog dataset has more delicate semantic features (i.e., robust features) compared to the other datasets, and it is more difficult for attackers to detect them during perturbation.

\begin{figure}[h]
  \centering
  \includegraphics[scale=0.46]{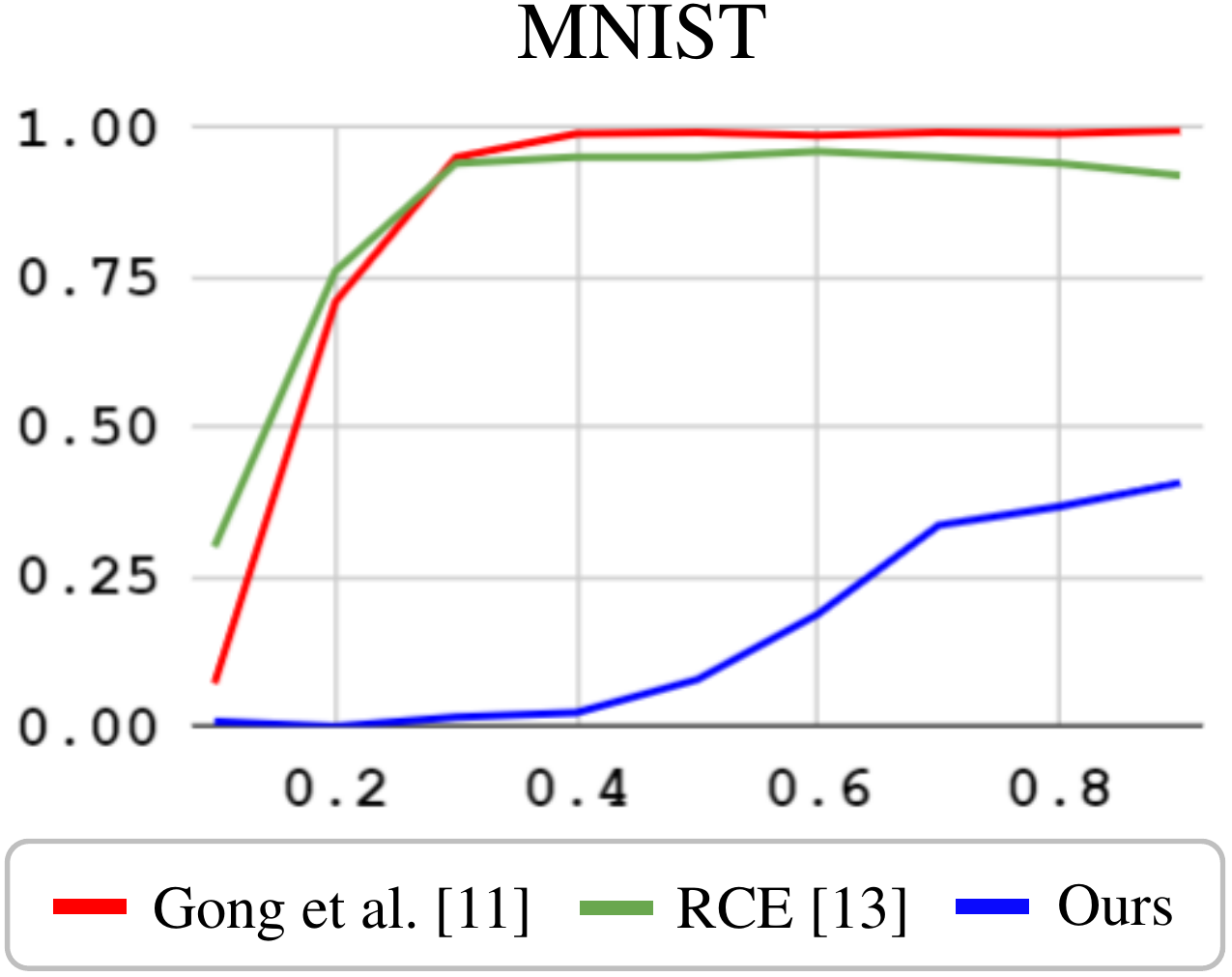}
    \caption{Baseline comparison on MIM\_W whitebox attack success ratio (C1, C2) along the $L_\infty$ distortion. Gong et al. and Ours are trained with MIM\_W attack}
  \label{fig:gong-rce-ours}
\end{figure}

In terms of the comparison of the state of the art, our defense mechanism shows clear semantic changes against the attacks.
For example Gong et al.'s auxiliarty classifier as a detector does not induce any robustness under adaptive whatbox attacks, and
Reverse cross entropy which tries to impose non-maximal entropy in training phase, also does not work in whitebox attack unlike with the report in the paper.
Actually it could not show significant robustness increase comapared to the Gong et al.'s approach in terms of the amount of distortion and semantic changes (see Figure~\ref{fig:gong-rce-ours}).

\subsection{Experiment environments}
We conduct our experiments on Ubuntu 16.04 machine with 4 GTX 1080 ti graphic cards and 64GB RAM installed.
We build our experiments with tensorflow version 1.12.0, on python version 3.6.8.


\end{document}